\documentclass[5p,times]{elsarticle}

\usepackage{amsmath,amsfonts}
\usepackage{algorithmic}
\usepackage{algorithm}
\usepackage{array}
\usepackage{subfig}
\usepackage{textcomp}
\usepackage{stfloats}
\usepackage{url}
\usepackage{verbatim}
\usepackage{graphicx}
\usepackage[hidelinks]{hyperref}
\usepackage{amssymb}
\usepackage{booktabs}
\usepackage{lscape}
\renewcommand{\citeauthor}[1]{#1 \textit{et al.}}
\usepackage{flushend}
\usepackage{caption}
\journal{Future Generation Computer Systems}

\begin{document}

\begin{frontmatter}

\title{EdgeSync: Accelerating Edge-Model Updates for Data Drift through Adaptive Continuous Learning} %% Article title
\author[inst1]{Runchu~Dong}
\ead{rcdong@stu.xjtu.edu.cn}
\author[inst1,inst2]{Peng~Zhao\corref{cor1}}
\ead{p.zhao@mail.xjtu.edu.cn}
\author[inst1]{Guiqin~Wang}
\ead{gqwang@stu.xjtu.edu.cn}
\author[inst1]{Nan~Qi}
\ead{qinan@stu.xjtu.edu.cn}
\author[inst1]{Jie~Lin}
\ead{jielinn@mail.xjtu.edu.cn}
\affiliation[inst1]{organization={School of Computer Science and Technology, Xi'an Jiaotong University},
            city={Xi'an},
            postcode={710049},
            country={China}}
\affiliation[inst2]{organization={National Engineering Laboratory for Big Data Analytics, Xi'an Jiaotong University},
            city={Xi'an},
            postcode={710049},
            country={China}}        
\cortext[cor1]{Corresponding author}        

%% Abstract
\begin{abstract}
Real-time video analytics systems typically deploy lightweight models on edge devices to reduce latency. However, the distribution of data features may change over time due to various factors such as changing lighting and weather conditions, leading to decreased model accuracy. Recent frameworks try to address this issue by leveraging remote servers to continuously train and adapt lightweight edge models using more complex models in the cloud. Despite these advancements, existing methods face two key challenges: first, the retraining process is compute-intensive, causing significant delays in model updates; second, the new model may not align well with the evolving data distribution of the current video stream. To address these challenges, we introduce EdgeSync, an efficient edge-model updating approach that enhances sample filtering by incorporating timeliness and inference results, thus ensuring training samples are more relevant to the current video content while reducing update delays. Additionally, EdgeSync features a dynamic training management module that optimizes the timing and sequencing of model updates to improve their timeliness. Evaluations on diverse and complex real-world datasets demonstrate that EdgeSync improves accuracy by approximately 3.4\% compared to existing methods and by about 10\% compared to traditional approaches.
\end{abstract}

\begin{highlights}
\item A more efficient edge-model updating approach that automatically and continuously adapts models to the scene with data drift.
\item A novel method for filtering video streaming samples that integrates timeliness and adaptability to eliminate unnecessary samples.
\item A continuous training manager that optimizes the training schedule and duration using both labeled and computed features.
\end{highlights}

%% Keywords

\end{frontmatter}

\section{Introduction}
Real-time video analytics has significant potential across a range of applications, including augmented reality, video surveillance, and traffic detection \cite{ananthanarayanan2017real}. Recent advancements in deep neural networks (DNNs) have significantly improved the performance of video analysis, with some models even surpassing human accuracy in certain scenarios \cite{he2016deep,chen2023diffusiondet,ma2023eos}. However, despite their accuracy, the complexity of DNN architectures and the large number of parameters contribute to a substantial computational burden \cite{dosovitskiy2020image}. As a result, these models face challenges in performing real-time analytics on resource-limited devices, such as mobile terminals and edge devices.

To meet the demands of real-time analysis on edge devices, deep neural networks with fewer parameters and shallower architectures are commonly deployed. However, in real-world scenarios, the distribution of video content can change over time due to factors such as lighting variations, crowd density, and weather conditions. This dynamic nature makes lightweight models more vulnerable to data drift \cite{smith2023closer}\cite{tan2019efficientnet}, resulting in difficulties maintaining the desired accuracy with a model trained offline. Recently, continuous learning has emerged as a feasible solution to enhance the adaptability of edge models. For instance, \citet{mullapudi2019online} demonstrated the effectiveness of this approach by proposing an online model distillation technique to train a low-cost student model on live video streaming. Ekya \cite{bhardwaj2022ekya} further refined this framework by performing data annotation in the cloud and jointly adjusting real-time inference and continuous learning-based model retraining on edge servers to maximize the overall accuracy through a scheduler.

\begin{figure}[!t]
\centering
\subfloat[Scene Change Example]{\includegraphics[width=3.3in]{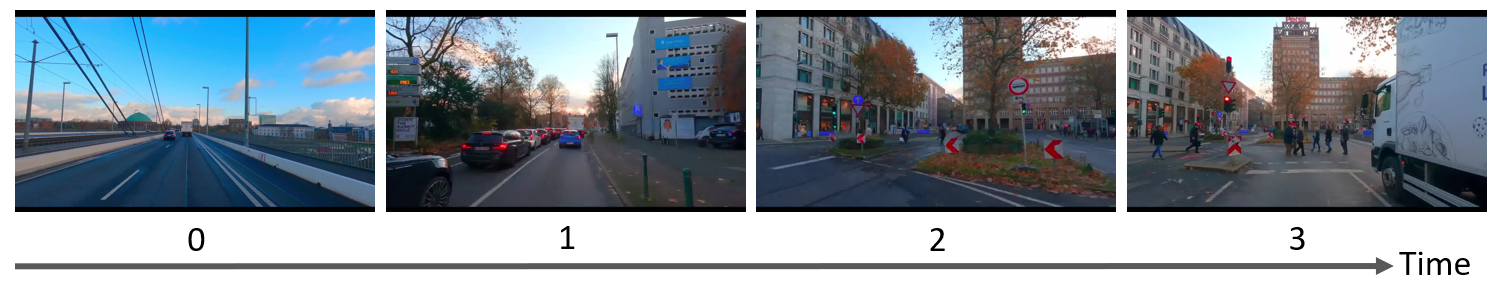}%
\label{fig_example_1}}
\hfil
\subfloat[Accuracy]{\includegraphics[width=1.65in]{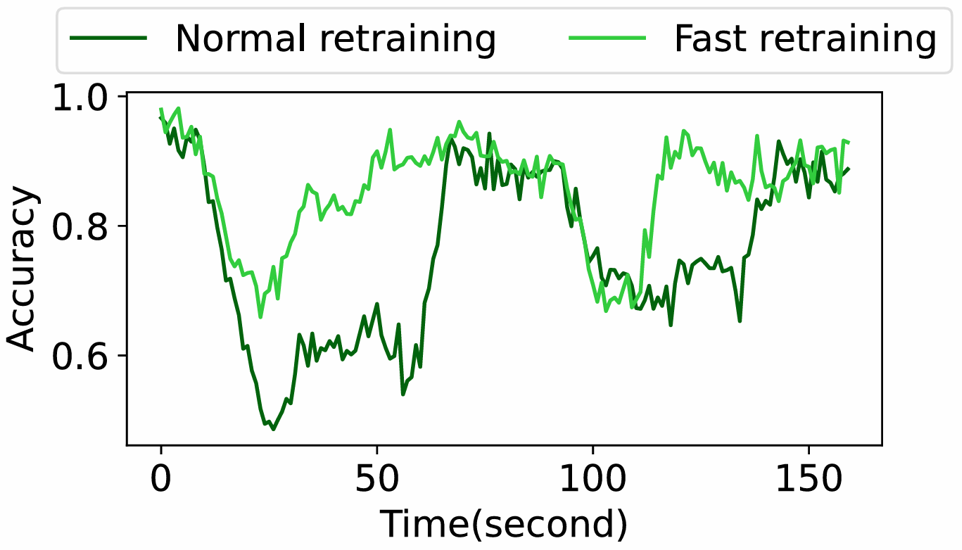}%
\label{fig_example_2}}
\subfloat[Class Distribution]{\includegraphics[width=1.65in]{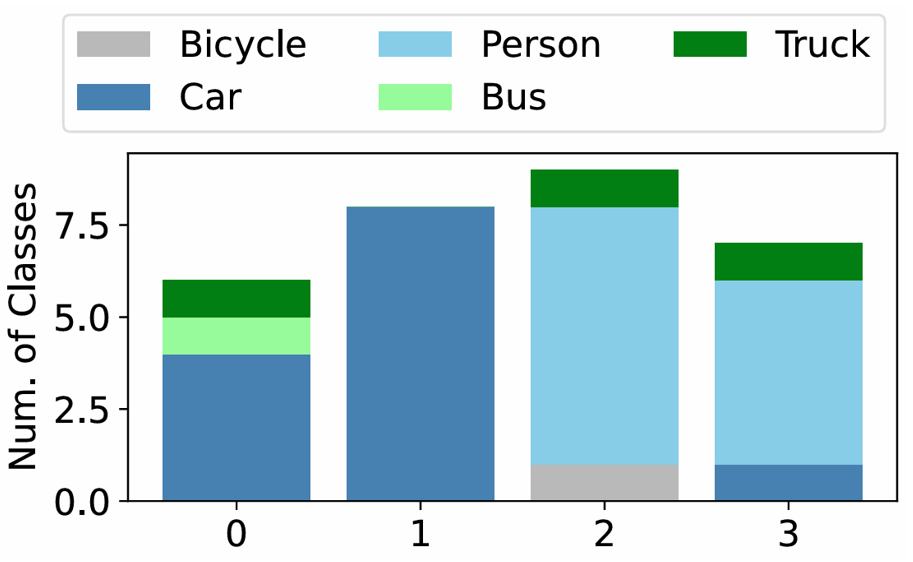}%
\label{fig_example_3}}
\caption{Changes of class distributions and accuracy when a car camera enters downtown of ideal fast retraining and normal retraining.}
\label{fig_example}
\end{figure}

Although existing research on continuous learning has the potential to enhance the inference accuracy of edge devices, several challenges remain. First, the update delay of new edge models remains substantial. Lightweight models at the edge continue to use outdated parameters for inference while retraining new ones. Existing approaches either upload all samples to the cloud for labeling and retraining or perform additional tasks, such as hyperparameter tuning, in the cloud. These processes consume significant time and contribute to increased response latency, particularly in the face of data drift, as illustrated in Fig.\ref{fig_example}. Second, existing methods typically use a static sampling rate and a fixed time interval for model updates. However, video streams exhibit varying degrees of change over time, and these methods fail to account for the differing impact of samples or the model's upper accuracy limit during each period. This limitation affects the quality of model retraining and reduces the generalization ability of the updated model.

To address these challenges, we introduce EdgeSync, a more efficient edge-model updating approach that automatically and continuously adapts models to the scene. EdgeSync comprises two primary modules: a sample filtering module and a continuous training management module. The sample filtering module operates in real time to reduce network bandwidth usage and enhance the quality of training samples. It selectively filters samples from the current video stream, based on the model's feature extraction capabilities and the importance of timeliness. The continuous training management module ensures high-quality model training while accelerating update speed, making accurate training handover decisions with minimal complexity. By integrating these modules, EdgeSync achieves faster model updates in real-world scenarios, enhancing the accuracy of single-camera inference and enabling the cloud to perform more camera model update tasks. The major contributions of this paper are summarized as follows:
\begin{itemize}
  \item First, we propose a novel method for filtering video streaming samples that integrates timeliness and adaptability to eliminate unnecessary samples. This approach enables flexible adjustment of the sample count based on network conditions, thereby enhancing the quality of training data in the cloud.
  \item Second, we introduce a continuous training manager that optimizes the training schedule and duration using both labeled and computed features. To further accelerate model updates, we utilize offline and online profiling to expedite hyperparameter selection.
  \item Third, extensive experiments were conducted to evaluate the performance of EdgeSync in complex scenarios. The results show that EdgeSync reduces network bandwidth consumption and increases the frequency of model updates, while outperforming baseline approaches in terms of overall accuracy.
\end{itemize}

The remainder of this paper is organized as follows: Section 2 reviews related work. Section 3 introduces the system architecture and details the sample filtering module and continuous training management module. Section 4 presents the evaluation results. Finally, concluding remarks are provided in Section 5.

\section{Related Work}

\subsection{Video Analytics Systems}
Real-time video analytics leverages computer vision algorithms to automatically analyze and interpret the content of video streams generated by one or more cameras. This enables the execution of complex tasks, such as target recognition and anomaly detection, while the video streams are being recorded and transmitted. To balance accuracy and processing speed, \citet{jiang2018chameleon} designed a controller that dynamically adjusts parameters within the video analysis system. \citet{wang2023edge} addresses the challenge of joint configuration tuning for both video parameters and serverless computation resources, proposing an algorithm that uses Markov approximation to optimize configurations based on the accuracy-cost trade-off. PacketGame \cite{yuan2023packetgame} enhances video processing by selectively filtering packets using a specially designed neural network before running the video decoder, thereby enabling more parallel video processing. Ekya \cite{bhardwaj2022ekya} enables simultaneous training and inference at the edge server, utilizing a thief scheduler to select appropriate parameters that balance both tasks. AMS \cite{khani2021real} and JIT \cite{mullapudi2019online} develop specialized lightweight DNNs to maintain accuracy under specific scenes. However, the challenge lies in the need to dynamically create new DNNs as video scenes change in order to adapt to new content. \citet{ruiz2025real} investigates real-time video analytics on low-power devices, with a focus on resource-efficient deployments. Our work aligns with these efforts but emphasizes an underexplored aspect: the update quality and speed of specialized lightweight DNNs. We aim to accelerate the model's adaptation to dynamic video content and enhance overall accuracy by improving the efficiency of model updates.

\subsection{Continuous Learning}
Continuous learning, also referred to as incremental learning, utilizes newly acquired data to adapt a model to new tasks while preserving previously learned fundamental concepts \cite{de2021continual}. Recent research has addressed the issue of catastrophic forgetting in the context of extended task sequences. For example, \citet{wang2022learning} introduces a meta-optimizer in SDML that dynamically adjusts the learning rate to mitigate forgetting during the learning process. The Gradient Projection Method (GPM) \cite{saha2020gradient} tackles this challenge by identifying critical gradient subspaces linked to previous tasks and preventing catastrophic forgetting through gradient steps orthogonal to these subspaces when learning new tasks. Additionally, some approaches focus on scenarios with limited training data. Notably, MAML \cite{finn2017model} fine-tunes model parameters using gradient descent for new tasks, while DeepBDC \cite{xiao2022few} improves image representation by assessing the discrepancy between joint characteristic functions of embedded features and their marginal products.

Existing approaches to continuous learning can be broadly categorized into three main strategies. The first category assigns distinct subsets of network parameters to each task \cite{yan2021dynamically}. The second category addresses forgetting in fixed-capacity models by applying structural regularization, which penalizes significant changes to parameters crucial for previous tasks \cite{lu2025continuous, hou2019learning}. The third category combats forgetting by either storing a subset of examples from past tasks in memory for rehearsal \cite{bang2021rainbow} or generating old data using generative models for pseudo-rehearsal \cite{shin2017continual}. In parallel, frameworks such as Kafka-ML \cite{carnero2024online} facilitate practical implementation in data streams, extending traditional pipelines to support online learning and continuous model updates through adaptive inference-training loops in both distributed and centralized architectures. Our approach aligns with the principles of continuous learning but introduces a novel framework. Rather than relying on traditional methods, we emphasize lightweight deep neural networks (DNNs) focused on recent video frames, leveraging spatial locality relations and training in real-time to enhance adaptability to dynamic video content.

\subsection{Unsupervised Adaptation Methods}

\begin{figure*}[t]
\centering
\includegraphics[width=6.21in]{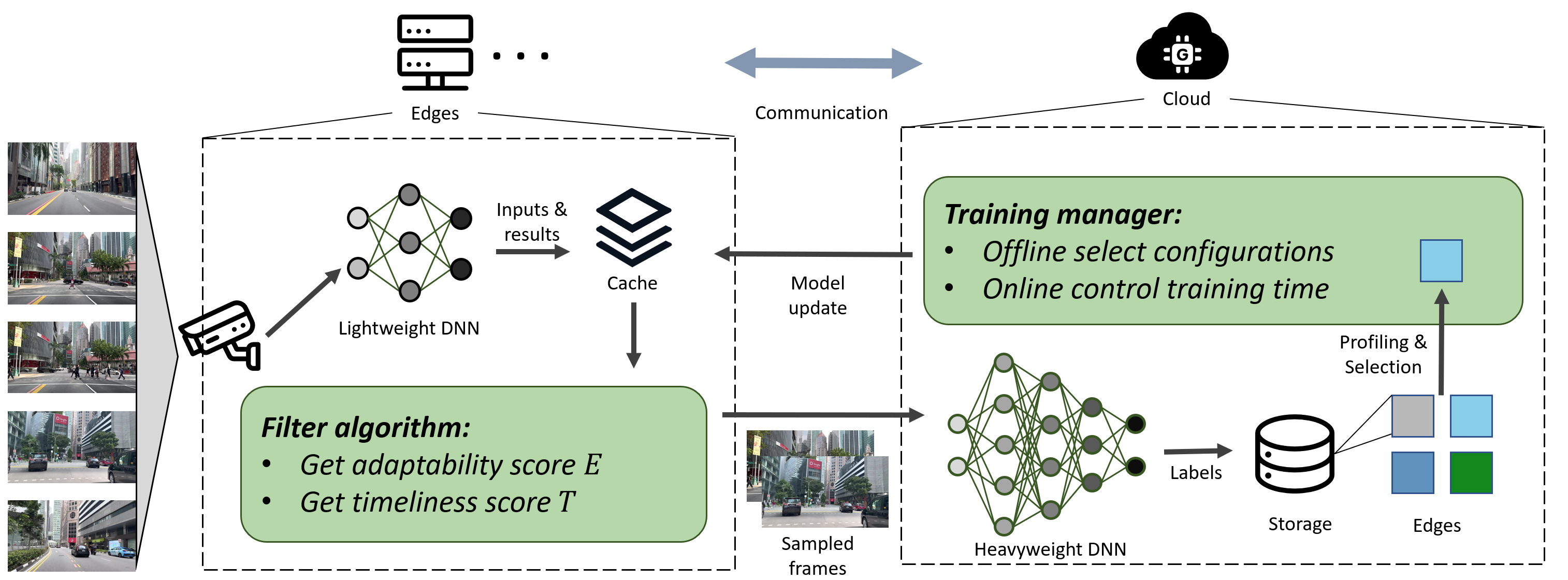}
\caption{Overall architecture of EdgeSync}
\label{fig_overview}
\end{figure*}

Unsupervised adaptation methods aim to enhance accuracy by addressing potential distribution shifts between the training and testing data. Early approaches to unsupervised domain adaptation involved fine-tuning models using both source and target domains \cite{long2016unsupervised, kang2019contrastive}. However, more recent methods focus exclusively on the target domain. For example, \citet{chidlovskii2016domain} proposed a method that requires only unlabeled test data. SHOT \cite{liang2020we} integrates entropy minimization with pseudo-labeling. \citet{fan2024test} introduces RTL, a technique that employs test-time linear regression to exploit the linear relationship between out-of-distribution (OOD) scores and network features for improved OOD detection. To reduce training time, Tent \cite{chen2022contrastive} optimizes the model’s confidence by minimizing test entropy, while LAME \cite{Boudiaf_2022_CVPR} adapts the model's output, rather than its parameters, by discouraging deviations from the predicted values. This approach helps identify the optimal set of latent assignments and minimizes the impact of hyperparameters on performance.

Additionally, several methods have been specifically tailored for video data. \citet{zeng2023exploring} proposed a test-time learning framework that utilizes motion cues in videos to improve the generalization ability of video classification models. \citet{lin2023video} introduced a test-time adaptation technique for video action recognition, which aligns training statistics with online estimates of target statistics and ensures prediction consistency across temporally augmented views of a video sample. Our approach employs a complex model hosted on a cloud server for supervised training, using labels generated by the model itself, making it more suitable for this framework.

\section{Method}

In this section, we present EdgeSync, our proposed method for real-time video analytics with adaptive continuous learning. We begin by describing the overall system architecture of EdgeSync, followed by a detailed explanation of its key components, including the sample filtering module and the adaptive continuous training manager.

\subsection{System Architecture}

Fig.\ref{fig_overview} illustrates the overall architecture of EdgeSync, a collaborative edge-cloud system designed for real-time video analytics. The system consists of multiple terminal cameras, several edge servers, and a centralized cloud server. Each edge device locally analyzes video streams from one or more cameras using a lightweight model. Meanwhile, the cloud server oversees the management of models across all edge devices, handling tasks such as model training and updates. The core components of both the edge and cloud servers are summarized as follows.

\textbf{Edge Server}: Each edge server deploys a lightweight model with a shallow architecture and a limited number of parameters to enable rapid local inference of real-time video streams. To maintain inference accuracy and mitigate the effects of data drift, the inference results and corresponding video frames are buffered in a local cache for online continuous learning. Since sending all frames directly to the cloud would cause significant network overhead and latency, the edge server must selectively upload samples. In this study, we propose a sample filtering module that evaluates the quality of samples over time and dynamically selects high-quality samples for training new models in the cloud. This module assesses sample quality from two perspectives. First, it calculates the entropy of the output confidence distribution derived from the current model on the edge; higher entropy indicates more informative updates for the current model. Second, it considers the temporal distribution, where the distance from the current time point reflects the similarity to the current sample period. The edge server sorts the samples based on these two criteria and selects the top several samples to send, along with their results and frames, to the cloud for retraining. Additionally, each edge server receives updated models from the cloud server for model updates.

\textbf{Cloud Server}: The cloud server utilizes sampled frames received from edge devices to dynamically train new models and then dispatches updates back to the edges. When data samples are received from an edge device, a continuous training manager in the cloud performs two main processes: labeling and retraining. To label new data samples, the cloud employs a complex, heavyweight model that leverages its high accuracy and extensive resources. This model generates highly accurate predictions, which are treated as ground truth labels to supervise the training of the lightweight edge models. The labeled data are stored in a buffer for future use. Before retraining, the cloud assesses the historical accuracy of each edge device to prioritize which model should be updated first. During retraining, the cloud uses pre-selected hyperparameters and an early-stopping mechanism to expedite the process. Once the model is updated, only the modified parameters, rather than the entire model, are sent back to the corresponding edge device. Additionally, the model and its parameters are stored in GPU memory to further accelerate updates and minimize delays caused by context switching between the GPU and main memory.

\subsection{Sample Filtering Module}
In practice, it is neither necessary nor efficient for edge devices to process all input frames simultaneously, as not all frames contribute equally to enhancing model retraining performance. To improve the quality of samples and accelerate the uploading process, we propose a sample filtering module that selectively uploads samples from the edge.

Before detailing the sample filtering module, we first define the relevant terms and assumptions. Within an update window $T$ on an edge server, the following elements are considered: (1) the model $f$ deployed on the edge server, characterized by parameter $\theta$; (2) the video frame $x_i$, where $i$ denotes the frame's sequence within the current window; (3) the inference result $y_i$ for frame $x_i$ using the current edge model $f$; and (4) cache $Y$, which stores inference results and the corresponding frames, and cache $M$, which stores the processed samples along with their computed quality scores. It is important to note that the size of the update window $T$ varies dynamically.

In the filtering module, we focus on two main aspects. First, we assess the adaptability of the edge model to the current sample. If the model accurately predicts the sample, retraining is unnecessary; however, poorly predicted samples require retraining. Second, we consider the timeliness of the sample. Since our goal is to improve the accuracy of the lightweight model by adapting it to local frames, it is crucial to select samples that represent the current video stream distribution. The filter must prioritize samples closer to the current timestamp while ensuring a sufficient amount of retraining data.

For the adaptability score $E\left(x;\theta\right)$, we avoid using the model's output confidence due to the poor calibration of modern neural networks \cite{DBLP:conf/icml/GuoPSW17}. Instead, we use the entropy of the model outputs to assess adaptability, as this method has proven effective in other domains:

\begin{equation} \label{eq:1}
E\left(x;\theta\right)=-f_\theta(y|x) \cdot \log{f_\theta(y|x)},
\end{equation}
where $x$ represents the current sample, and $y$ is the result predicted by model $f$ with parameter $\theta$. A higher adaptability score suggests that the model has lower certainty about the current sample. To calculate the timeliness score for sample $x$, denoted as $T(x)$, we use the following formula to assess its importance:
\begin{equation} \label{eq:2}
T(x)=1/(1+\exp(-t_x/T_w)),
\end{equation}
where $T_w$ is the current window size, which is the duration from the last training session to the present, and $t_x$ denotes the relative timestamp of sample $x$, indicating its timeliness in relation to the current timestamp. For the most recent sample, $t_x$ is typically initialized to 0, while older samples will generally have a larger relative timestamp.

Finally, the overall quality score of the sample $x$, denoted as $Q(x)$, is calculated as a weighted average of the adaptability score and the timeliness score, as follows:
\begin{equation} \label{eq:3}
Q\left(x\right)=\alpha E\left(x;\theta\right)+\beta T(x),
\end{equation}
where $\alpha$ and $\beta$ are parameters used to balance adaptability and timeliness.

\begin{algorithm}[!t]
	\caption{Sample Filtering Process}\label{alg:cap}
	\textbf{Input:} Cache $Y \gets \{(x_1, y_1), (x_2, y_2), ..., (x_n, y_n)\}$.\\
	\textbf{Output:} Filtered sample list result $M$.
	
	\begin{algorithmic}[1]
	\STATE \textbf{Initialization:} update window size $T$; percentage of uploaded samples $k$; $\alpha$; $\beta$; Cache $M \gets \emptyset$
    \FOR {$(i, x, y) \in Y$}
    \STATE Calculate adaptability score $E(x;\Theta)$ via Eq.(\ref{eq:1});
    \STATE Calculate timeliness score $T(x)$ via Eq.(\ref{eq:2});
    \STATE Calculate overall quality score $Q(x)$ via Eq.(\ref{eq:3});
    \STATE Append $(x, y, Q)$ to list $M$; 
    \ENDFOR
    \STATE $M \gets sort(M)$ sort element in $M$ by overall score;
    \STATE $M \gets$ get top-$k$ samples from sorted list $M$; 
    \STATE send $M$ to the cloud
    \end{algorithmic}
\end{algorithm}

Based on the sample evaluation process described, we have designed the sample filtering module, detailed in Algorithm \ref{alg:cap}. The pseudo code outlines the procedure for filtering video frames and their corresponding inference results to optimize the data sent to a cloud server for further retraining. Initially, the algorithm defines the input cache $Y$, which contains pairs of video frames $x$ and their corresponding inference results $y$. Each entry in $Y$ represents a sample consisting of a frame and its prediction result. The code then initializes several parameters, including the update window size $T$, the percentage $k$ of top samples to upload, and the scoring weights $\alpha$ and $\beta$. An empty cache $M$ is created to store processed samples. The algorithm iterates through each sample in the cache, computing three key metrics for each frame: the adaptability score $E(x;\Theta)$ for frame $x$ using the model's parameters $\Theta$, as specified by Eq.(\ref{eq:1}); the timeliness score $T(x)$ for the sample $x$, based on Eq.(\ref{eq:2}); and the overall quality score $Q(x)$, which combines the adaptability and timeliness scores according to Eq.(\ref{eq:3}). Then, each frame, along with its result and quality score, is added to the list $M$. After processing all samples, the list $M$ is sorted in descending order by the overall quality score. From this sorted list, the top-$k$ samples—determined by the specified percentage—are selected. Finally, the selected samples are packaged and sent to the cloud server for further model retraining. This approach ensures that only the most relevant and high-quality samples are transmitted, optimizing the efficiency of data processing and model updates.

\subsection{Continuous Training Manager Module}

Upon receiving the filtered samples from the edge devices, the training manager module in the cloud will continuously learn knowledge from these samples to enhance the model's performance. Specifically, the continuous training manager module dynamically updates the model parameters and adjusts the training sequence to optimize overall accuracy.

\subsubsection{Model selection for continuous training}

Different edge devices may experience varying degrees of data drift in their video content. For example, cameras monitoring road conditions are placed at various locations throughout a city, each facing different patterns of change due to their unique geographical contexts. To determine which model should be prioritized for retraining, we use a historical window-based detection method, which evaluates the urgency of edge models by assessing the model's inference accuracy from the last training session to the current period.

In particular, the cloud receives samples along with their corresponding inference results transmitted from the edge. We propose to build up profiling problem based on the model inference result level rather than the sample feature level, as this approach provides a more robust signal for detecting scene changes. This method also avoids the additional computational overhead associated with approaches that introduce a feature extractor at the cloud. We then employ a history window-based detection strategy to process these inference results. Specifically, in the cloud, samples are first sent to a high-accuracy, complex model to generate pseudo-labels. The inference results are then compared to these labels, producing a binary variable $acc \in \{0, 1\}$ to indicate the correctness of each sample. These values are appended to the list $W=\{acc_1,acc_2,\ldots,acc_n\}$ in cloud storage, where $n$ represents the list's capacity. It is important to note that each edge will have a different $W$ in the cloud. The retraining management module uses $W$ to assess the urgency degree $d$ for edge:
\begin{align}
    d &= \sum_{i=0}^{m}{\left({wa}_0-{wa}_i\right)\cdot(1/(1+e^{-i/tm})\cdot m)} \label{eq:4} \\
    wa_i &= \sum_{j=l \cdot i+1}^{(l+1) \cdot i}{acc_j} \label{eq:5}
\end{align}
where $l$ is the length of each batch $wa$, which sums up $acc$ variables, $m$ is the number of batch $wa$ in list $W$.

We also consider temporally significant differences in window $W_i$ via an exponential decay weight, which is similar to the sample filtering module. The retrain manager continuously removes outdated samples when the number of $acc$ exceeds list's capability.

\begin{algorithm}[!t]
\caption{Training Manager Algorithm}\label{alg:train}
\textbf{Input:} samples and inference results received from edge $e$: $M_e=\{(x_{1e}, y_{1e}, 1), (x_{2e}, y_{2e}, 2),\ldots , (x_{re}, y_{re}, r)\}$, hyperparameters $h$ obtained in offline phase, early stop threshold $k$, complex model $f$, list size $n$, and batch size $m$.\\
// $x_{ie}$ is the input sample from edge node $i$, $y_{ie}$ is the ground truth label, and $r$ denotes the edge node ID.\\
\textbf{Output:} Retrained new model
\begin{algorithmic}[1]
\FOR{$e \in \text{edges}$}
    \FOR{$(x, y, i) \in M_e$}
        \STATE $label \gets f(x)$; \\// Model inference on the sample x and get its lable
        \STATE $acc \gets (label == y)$; \\// Compute the accuracy for this sample
        \STATE $\text{append } (acc, i) \text{ to } e.\text{bank}$;
    \ENDFOR
    \IF{$\text{len}(e.\text{bank}) > n$}
        \STATE remove excessive elements from $e.\text{bank}$; \\// Ensure the list size does not exceed $n$
    \ENDIF
    \STATE \text{divide } $e.\text{bank}$ \text{into } $m$ \text{batches};
    \STATE $d_e \gets \text{Score}(e.\text{bank}, m)$;  \\// Calculate urgency degree score for edge $e$ via Eq. (\ref{eq:4})
\ENDFOR
\STATE $e \gets \arg\max d_e$;  \\// Select the edge with highest urgency score
\STATE $epoch \gets 0$, $max\_epoch \gets 0$;
\STATE $start\_time \gets current\_time()$,  $max\_evaluation \gets 0$;
\WHILE{$epoch-max\_epoch\leq k$ \textbf{and} $current\_time()-start\_time < max\_time$}
    \STATE \text{train model for edge} $e$ \text{ using hyperparameters } $h$;
    \STATE $evaluation \gets e.\text{eval}()$;  \\// Evaluate model performance
    \IF{$evaluation > max\_evaluation$}
        \STATE $max\_evaluation\gets evaluation$, $max\_epoc \gets epoch$;
    \ENDIF
    \STATE $epoch \gets epoch + 1$;
\ENDWHILE
\end{algorithmic}
\end{algorithm}

\subsubsection{Dynamic updating frequency for adaptive training}
Ideally, when a video stream undergoes significant changes, the training module should respond quickly by expediting the scene adaptation process. This requires allocating additional computational resources for retraining models deployed on edge devices and shortening the feedback loop for model updates to minimize the update window. While the model analysis mechanism discussed earlier can quantify content change intensity to optimize time allocation strategies, it has the inherent limitation of not directly controlling model training duration. To address this, the cloud must be equipped with a dynamic retraining management module that can adaptively adjust training schedules based on real-time demands.  

The existing research, such as that by Ekya, has highlighted the significant impact of hyperparameter training on model accuracy, underscoring the necessity for more cautious parameter selection strategies in practical applications. However, applying Ekya's micro-analytical approach to our framework presents two challenges. First, our method's operational paradigm requires more frequent model updates, substantially increasing both computational time and resource investment for online analysis during each update cycle. This can reduce overall operational efficiency. Second, Ekya's approach to parallel training and inference execution necessitates the management of numerous interrelated hyperparameters, whereas our framework focuses on the targeted optimization of a key subset of parameters during training. While this reduces the complexity of parameter regulation, it also introduces discrepancies in parameter scheduling between the two approaches. Through comparative experiments and analysis, we conclude that offline analysis methods are better suited to the characteristics of this architecture. Offline methods allow for centralized optimization of retraining durations for active edge models and effectively compress the complete update cycle through strategic time-resource allocation, thereby enhancing overall system performance while maintaining response speed.  

\paragraph{\textbf{Offline Phase}}

The proposed method utilizes a two-step strategy in the offline phase. First, Bayesian Hyperparameter Optimization (BHO) is employed to derive initial hyperparameter configurations from a variety of video data samples. Second, mini-batch iteration is applied to refine and update these initial parameters, ultimately producing the final offline hyperparameters.  

In the offline phase, a video dataset encompassing multiple scenarios is constructed to ensure the samples are sufficiently representative. For each video sample in the dataset, the BHO algorithm autonomously identifies the optimal combination of hyperparameters. As a global optimization technique based on probabilistic surrogate models, the core of BHO involves using a surrogate model, typically a Gaussian Process (GP), to approximate the relationship between hyperparameter configurations and target performance.  

In the GP model, the target function \( f(\theta) \) is assumed to follow the distribution:  
\[
f(\theta) \sim \mathcal{GP}(\mu(\theta), k(\theta, \theta'))
\]  
where \( \mu(\theta) \) represents the mean function, and \( k(\theta, \theta') \) is the covariance kernel function describing the correlation between any two points. Using the observed data \( \{(\theta_i, f(\theta_i))\}_{i=1}^n \), we can predict the values of the target function at new hyperparameter configurations \( \theta \), providing both the predicted mean \( \mu(\theta) \) and the associated uncertainty (variance) \( \sigma^2(\theta) \).  

Based on the surrogate model, the BHO method utilizes an acquisition function to identify the next hyperparameter point for evaluation. A commonly employed acquisition function is Expected Improvement (EI), which is defined as follows:  
\[
\text{EI}(\theta) = \mathbb{E}\left[\max\{f(\theta) - f^+,\, 0\}\right]
\]  
where \( f^+ \) represents the best observed performance so far. The EI function strikes a balance between exploring uncertain regions (high uncertainty) and exploiting well-performing regions (high predicted mean). At each iteration, the next hyperparameter configuration to evaluate is chosen by solving:  
\[
\theta^* = \arg\max_{\theta} \text{EI}(\theta)
\]  
The new observation \( (\theta^*, f(\theta^*)) \) is then incorporated into the dataset to update the Gaussian process model, enabling the algorithm to efficiently identify high-quality hyperparameter combinations within a limited number of iterations. Finally, the global initial hyperparameter baseline \( h_0 \) is computed as the arithmetic mean of the hyperparameters obtained from all video samples. 

After obtaining the initial baseline \( h_0 \), the algorithm proceeds to the iterative optimization stage to refine hyperparameter configurations. The process unfolds as follows: First, temporally continuous video segments are randomly sampled from each video to form training batches, ensuring both temporal coherence and continuity of the data. Second, for each training batch, forward and backward propagation are executed, and gradient information is leveraged to dynamically adjust hyperparameter configurations. Following each iteration, the retraining management module computes and records the performance improvement on the validation set (e.g., accuracy metrics). The optimization process terminates when performance improvement falls below a preset threshold \( \varepsilon \) for \( N \) consecutive iterations, thereby preventing overfitting or ineffective updates.  

This strategy of using BHO for initial hyperparameter search followed by mini-batch iterative optimization effectively combines a global and efficient search strategy. It also dynamically improves model stability and performance through adaptive adjustments, thereby reducing the computational burden during the subsequent online phase.

\paragraph{\textbf{Online Phase}}

In the online phase, the cloud executes dynamic epoch adjustment (\(k\)) through a dual safeguard mechanism. While all hyperparameters remain fixed at their initial values obtained from the BHO, the system first applies progressive early stopping: if the validation loss reduction \( \Delta L \) falls below a threshold \( \delta \) for \(k\) consecutive cycles, training automatically terminates. This adaptive epoch adjustment prioritizes convergence efficiency without altering hyperparameters. Simultaneously, a time-constrained termination protocol enforces a global maximum training duration \(T_{max}\). By decoupling hyperparameter fixation from dynamic epoch adjustment, the method achieves two-tier optimization: conserving computational resources through fixed parameters and enabling flexible training control via adaptive \(k\) adjustment.

Notably, a hierarchical parameter freezing strategy is implemented throughout the training process: the weights of the backbone network (e.g., ResNet) and shallow feature extraction layers remain fixed, while only the weights of the final fully connected and classification prediction layers are fine-tuned. This approach is grounded in the feature learning characteristics of deep neural networks: shallow layers capture generic visual features (such as edge textures and color distributions), while deeper layers gradually learn task-specific representations (such as object geometric structures and semantic correlations), as validated in several studies \cite{bau2017network}\cite{guo2019spottune}.

\subsubsection{Edge model updating}
After completing the training process, the training manager module returns the newly trained model to the relevant edge device. It is important to note that only the updated parameters, rather than the entire edge model, are transmitted over the network. Additionally, to accelerate model updates and minimize context-switching delays between the GPU and memory, both the model and frozen parameters are permanently stored in the edge device's GPU memory.

\section{Experiments}
In this section, we evaluate the performance of the proposed method from various perspectives. Specifically, we first present the experimental setup, including the datasets, baselines, and implementation details. Subsequently, we present and analyze the experimental results.

\subsection{Experimental setup}

\subsubsection{Datasets}

We evaluate EdgeSync using 22 video segments sourced from YouTube, each with a frame rate of 30 frames per second. To incorporate meaningful data drifts, we select videos approximately 20 minutes long, resulting in a total video length of about 7 hours. These videos exhibit varying light conditions (day and night), weather conditions (sunny, rainy, snowy), and movement speeds (walking, driving), thereby presenting a range of data drift scenarios with different levels of complexity. To convert the videos into images, we randomly select two frames per second and label them in chronological order. For experiments involving multiple cameras, the data is evenly distributed across the cameras, and each camera is assigned unique scenes to avoid repetition. Notably, we do not use common public datasets like Cityscapes\cite{Cordts_2016_CVPR}, as our goal is to have video data containing multiple complex scenes while maintaining temporal consistency and ensuring each video is of sufficient length.

\subsubsection{Compared Approaches}
To verify the effectiveness of EdgeSync, we compare it with the following baseline methods:
\begin{itemize}
\item No Adaptation: This approach executes the pre-trained model on the edge device without any further adaptation.

\item One-Time Adaptation (Adap.)\cite{rebuffi2017icarl}: Serving as a baseline for scenarios without continuous adaptation, this method involves fine-tuning the edge models using the first 100 seconds of samples at the beginning of each video, similar to transfer learning. After this initial adjustment, no further updates to the parameters are made.

\item AMS\cite{khani2021real}: The architecture of AMS is similar to EdgeSync, featuring repeated training and updates of the edge model with cloud support. However, AMS employs a fixed number of epochs to train the edge model and updates only a small fraction of its parameters. In our experiments, we update the parameters of the last feature layer and the final classification layer to ensure a fair comparison.

\item Ekya\cite{bhardwaj2022ekya}: Ekya performs both inference and retraining tasks on edge servers using a fixed time window and a micro-profiler to determine resource allocation for each window. In our implementation, the retraining task and micro-profiler are hosted in the cloud, determining training hyperparameters for the current window, which is set to 200 seconds. The micro-profiler considers factors such as the number of layers to retrain, the fraction of data to use between retraining windows, momentum in SGD, and weight decay. We set the fraction of training data to 20\% and the early termination epoch to 5.
\end{itemize}

\subsubsection{Models}
For edge devices, we use MobileNetV2\cite{Sandler_2018_CVPR}, which provides real-time inference speeds of 30 frames per second, even on devices with lower computational power like the NVIDIA Jetson Nano and Jetson TX2. In contrast, on the cloud server, we use a high-performance model, ResNeXt101\cite{wang2019elastic}, to generate accurate ground truth labels. Both models for the classification task are pretrained on ImageNet datasets\cite{deng2009imagenet}.

\subsubsection{Implementation Details}
We use the NVIDIA GeForce RTX 2080Ti GPU as our edge device. To simulate a real-world scenario, we calculate the time interval between two model updates and then adjust the actual inference speed of the edge device based on the FPS of the Jetson Nano. All evaluations are conducted using a single NVIDIA Tesla V100 GPU in the cloud. The models on both the edge and cloud are implemented in Python 3.8 and PyTorch 2.0 with CUDA 11.7. On the edge device, we apply Algorithm 1 with parameters $k = 0.7$, indicating that 30\% of samples are filtered in the current window, and set $\alpha = 1.0$ and $\beta = 1.0$. In Algorithm 2, we configure the patience parameter to 5, with a memory bank capacity $N$ of 90 and a segment size $m$ of 10.

\subsection{Results and Discussions}
\subsubsection{Comparison to Baselines}
We first compare the end-to-end accuracy of EdgeSync with that of baseline methods, using seven concurrent cameras replaying videos from our dataset. To evaluate accuracy, we perform a classification task, comparing the edge device's inference results with labels generated by the teacher model for video frames. Accuracy is defined as the ratio of correct predictions to the total number of cases across six categories: people, bicycles, cars, motorcycles, buses, and trucks. 

\begin{table}[h]
\caption{Accuracy and bandwidth consumption of different methods}
\centering
\begin{tabular}{@{}llll@{}}
\toprule
Method              & Accuracy & Upload & Download  \\ \midrule
No Adaptation       & 62.40\%  & 0.0 Kbps           & 0.0 Kbps             \\
One-Time Adap.       & 63.96\%  & 42.47 Kbps         & 11.13 Kbps           \\
AMS                 & 68.87\%  & 254.84 Kbps        & 352.74 Kbps          \\
Ekya                & 68.70\%  & 254.84 Kbps        & 358.4 Kbps           \\
EdgeSync            & 72.09\%  & 178.39 Kbps        & 1836 Kbps            \\ \bottomrule
\end{tabular}
\label{table_e2e}
\end{table}

Table \ref{table_e2e} presents the overall accuracy of five methods. The findings indicate that continual training of the edge model leads to significant improvements in accuracy. Among the methods, EdgeSync achieves the highest performance, with an 8\% increase in accuracy compared to the No Adaptation method. The No Adaptation method exhibits the lowest accuracy, primarily due to its inability to adapt to specific scenes. While the One-Time Adaptation method shows a slight improvement over No Adaptation, it remains suboptimal as it cannot accommodate future video content. AMS, which consistently uses samples from the most recent time period for training, helps the edge model maintain a stable level of accuracy, resulting in a 4.9\% improvement over One-Time Adaptation. Ekya's accuracy is similar to that of AMS, though it employs a different approach. While AMS dedicates all cloud resources to model training and sample labeling, treating each edge model uniformly, Ekya adopts a dynamic strategy. It considers the impact of current window configurations on accuracy during continuous training and selects more appropriate configurations, even if this slightly delays the start of the window. This approach is particularly beneficial when the cloud manages numerous edge tasks, as the time spent on configuration selection becomes a more significant portion of the total time. However, this also limits the overall accuracy improvement of the edge model.

\begin{table*}[h]
\caption{Time cost of different methods within a cycle}
\centering
\begin{tabular}{@{}llllll@{}}
\toprule
Method    & Label time & Retraining time & Model profiling time & Network communication time & Total time     \\ \midrule
AMS       & 43.83 s    & 66.25 s         & 0.1 ms               & 3.52 s                     & 113.6 s        \\
Ekya      & 33.2 s     & 58.49 s         & 7.84 s               & 3.52 s                     & 103.05 s       \\
EdgeSync* & 9.67 s     & 13.72 s         & 0.1 ms               & 3.52 s                     & 26.91 s        \\
EdgeSync  & 9.02 s     & 8.77 s          & 1 ms                 & 3.52 s                     & 21.311 s       \\ \bottomrule
\end{tabular} 
\label{table_time}
\end{table*}

\subsubsection{Time Cost} 
Table \ref{table_time} presents the average time required to update a model using various methods within a single update window. EdgeSync* denotes the EdgeSync method without the retraining manager module. The time cost of each retraining process consists of four components: labeling, edge model profiling, retraining, and edge-cloud communication. EdgeSync has the lowest average total time compared to other methods. The time distribution for these four components in EdgeSync is 42.32\%, 41.15\%, 0.01\%, and 16.52\%, respectively. As previously discussed, minimizing model update time helps prevent the retrained model from becoming outdated, thereby enhancing adaptation to complex environments. Without the retraining manager module, EdgeSync experiences a 26.27\% increase in single update time, due to increased sample labeling time and extended training duration. AMS requires 43.83 seconds for labeling and 66.25 seconds for retraining. While AMS’s longer training time horizon leads to better generalization of the edge model, its adaptability diminishes when the edge video scene changes rapidly. Ekya employs a time-consuming heuristic that evaluates each candidate pair to determine which will most improve accuracy. According to Table \ref{table_time}, profiling for each edge takes 7.84 seconds, accounting for 26\% of the total runtime in our experiments, which adds significant computational overhead. Additionally, the model in Ekya can only be updated once within a window, limiting its effectiveness for edges that require frequent updates.

\subsubsection{Impact of the Number of Cameras} 
\begin{figure}[t]
\centering
\includegraphics[width=0.4\textwidth, keepaspectratio]{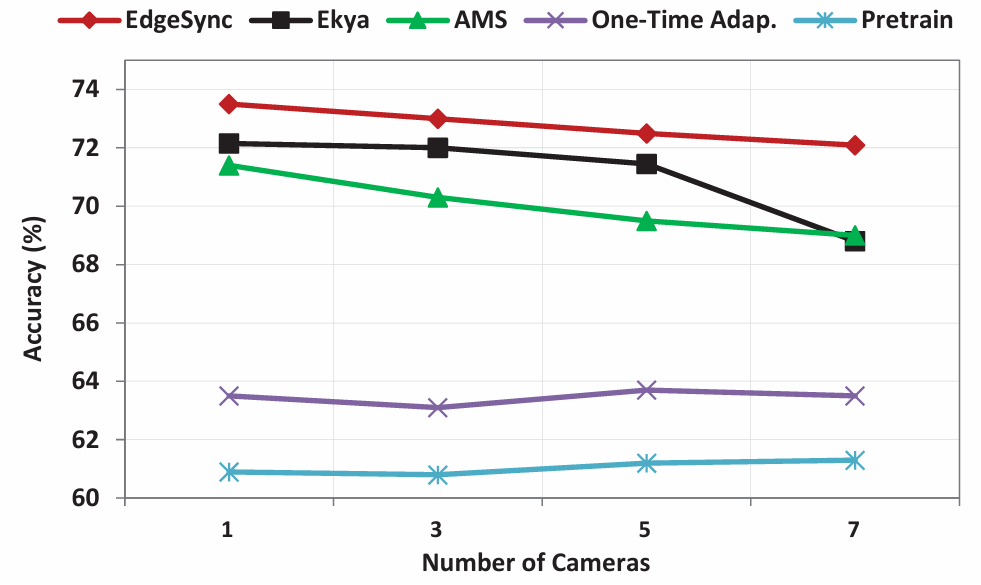}
\caption{Accuracy with varying numbers of cameras}
\label{fig_num}
\end{figure}

Fig.\ref{fig_num} illustrates the impact of the number of cameras on overall accuracy. The No Adaptation and One-Time Adaptation methods do not involve continuous tuning during task execution and therefore do not compete for cloud resources. As a result, the accuracy of these methods varies only slightly, primarily due to differences in the video data captured by each camera. In contrast, the accuracy of AMS, Ekya, and EdgeSync decreases as the number of cameras increases. This decline is due to the reduced retraining frequency for each lightweight model and the overall decrease in available training time, which diminishes the model’s adaptability to current video content at the edge. Compared to AMS and Ekya, EdgeSync’s accuracy decreases more gradually—1.2\% less than AMS and 2.2\% less than Ekya—as the number of cameras increases. This distinction arises from EdgeSync's adaptive strategy, where it autonomously determines when to stop the training process. This approach allows for more frequent model updates by shortening training time while still ensuring sufficient training duration. Furthermore, EdgeSync utilizes retraining samples with the most informative gradients, which improves the current lightweight model's adaptability to the prevailing video distributions.

\subsubsection{Offline hyper-parameters profiling performance.}  
\begin{figure}[ht]
\centering
\subfloat[Verification accuracy]{\includegraphics[width=1.55in, height=1.1in]{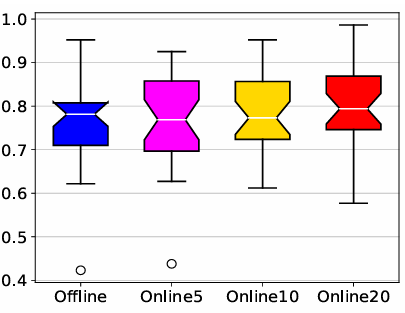}%
\label{fig_first_case}}
\hfil
\subfloat[Hyper-parameters profiling cost]{\includegraphics[width=1.55in, height=1.14in]{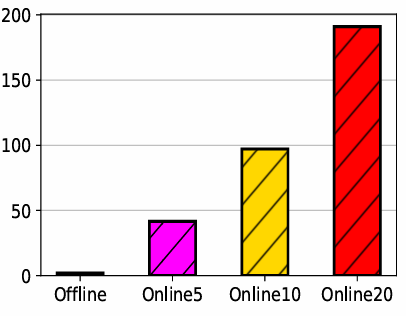}%
\label{fig_second_case}}
\caption{Impact of hyper-parameters profiling}
\label{fig_ver}
\end{figure}

To demonstrate the effectiveness of the offline hyper-parameters profiling method, we compare it to the online profiling method. In this experiment, the online profiling employs the Tree-structured Parzen Estimators (TPE) algorithm\cite{bergstra2011algorithms} to optimize the learning rate, momentum, and L2 penalty term, with the number of samples parameter set to 5, 10, and 20. This parameter specifies the number of iterations for the search algorithm. Fig.\ref{fig_ver} illustrates the accuracy results and time expenditure of online dynamic profiling with offline fixed hyper-parameter profiling. We find that with a smaller number of samples, online profiling performs worse than offline profiling, requiring a longer search process to achieve better hyper-parameters. When the number of samples is set to 10, the accuracy from online profiling is only slightly higher (0.1\%) than that of offline profiling over 31 consecutive windows. However, each online profiling session incurs an additional 100 seconds, significantly impacting the speed of model updates.

\subsubsection{Ablation Study} 
In this experiment, we systematically evaluated the impact of each module by incorporating them individually to determine their effect on accuracy. Specifically, EdgeSync/STF refers to EdgeSync without model selection and sample filtering at the edges, with a fixed update time equivalent to the window averaging time used in EdgeSync for this experiment. EdgeSync/TF denotes EdgeSync without dynamic training time adjustments and sample filtering at the edges. EdgeSync/F represents EdgeSync without sample filtering at the edges.

As shown in Table \ref{table_abEdgeSync}, EdgeSync/STF exhibits a decrease in accuracy of approximately 3.6\% compared to EdgeSync and a slight decrease of 0.4\% compared to AMS. This suggests that reducing the model update time alone, as in AMS, can compromise accuracy, likely due to overfitting, particularly when there is a shift in frame distribution. EdgeSync/TF achieves about 0.5\% higher accuracy than EdgeSync/STF, highlighting the importance of profiling edge models to select the most impacted ones before retraining. EdgeSync/F shows an improvement of approximately 1\% over EdgeSync/TF, demonstrating the effectiveness of dynamically adjusting the update time during runtime. However, EdgeSync/F still shows a decrease of about 1\% compared to EdgeSync, as it does not account for the varying impact of different samples on the current model.

\begin{table*}
\caption{Ablation performance of different modules of EdgeSync}
\centering
\begin{tabular}{@{}cccccc@{}}
\toprule
Method     & EdgeSync   & AMS      & EdgeSync/STF     & EdgeSync/TF    & EdgeSync/F      \\  \midrule
Accuracy   & 0.7210     & 0.6880   & 0.6840           & 0.6950         & 0.7108          \\   \bottomrule
\end{tabular}
\label{table_abEdgeSync}
\end{table*}

\subsubsection{Influence of upload percentage $k$.} 
\begin{figure}[t]
\centering
\includegraphics[width=0.4\textwidth, keepaspectratio]{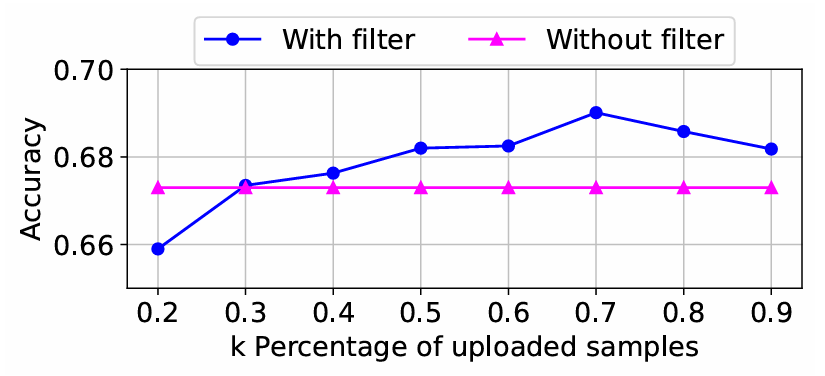}
\caption{Sensitivity analysis of filter percentage parameters}
\label{fig_k}
\end{figure}

Fig.\ref{fig_k} illustrates the overall accuracy and training time associated with various sample filtering percentages. To address potential accuracy inflation from increased model update frequency, we conducted experiments with a fixed model update interval of 100 seconds, similar to AMS with a filter sampler. The results show that baseline accuracy (with unfiltered samples) is comparable to accuracy achieved with 30\% sample filtering. However, selecting a small percentage of samples, such as 20\%, leads to reduced performance due to insufficient training data, which is crucial for effective deep learning. As the amount of training data increases, model performance improves, reaching peak accuracy at a filtering ratio of 0.7. Beyond this ratio, accuracy decreases, likely due to the inclusion of some disruptive samples that diminish training effectiveness, particularly given the limited capacity of lightweight models. This highlights that sample filtering not only shortens individual model training times but also promotes training consistency, thereby enhancing overall prediction accuracy.

\section{Conclusion}
In this paper, we present EdgeSync, a novel approach designed to enhance the accuracy and efficiency of edge-based video analysis systems. By integrating dynamic sample filtering and adaptive model updating techniques, EdgeSync addresses key challenges in managing data drift and optimizing model performance on edge devices. Our experiments demonstrate that EdgeSync significantly outperforms baseline methods. The adaptive nature of EdgeSync, combined with its ability to autonomously manage model updates and sample filtering, results in improved accuracy and efficiency. Specifically, EdgeSync achieves up to an 8\% increase in accuracy compared to the No Adaptation method and shows a more gradual decline in performance as the number of cameras increases. Additionally, the study demonstrates the advantages of using offline hyper-parameter profiling over online methods, emphasizing the reduction in computational overhead and the ability to maintain high accuracy with fewer resources. The effectiveness of each module within EdgeSync further underscores the importance of tailored model adaptation and dynamic training strategies.

\bibliographystyle{elsarticle-num-names} 
\bibliography{cas-refs}

\end{document}